\documentclass{article}

\usepackage[preprint]{neurips_2025}

\usepackage[utf8]{inputenc} 
\usepackage[T1]{fontenc}    
\usepackage{hyperref}       
\usepackage{url}            
\usepackage{booktabs}       
\usepackage{amsfonts}       
\usepackage{nicefrac}       
\usepackage{microtype}      
\usepackage{xcolor}         

\usepackage{lipsum}
\usepackage{fancyhdr}       
\usepackage{graphicx}       
\graphicspath{{media/}}     
\usepackage{amsmath}      
\usepackage{algorithm}       
\usepackage{algpseudocode}   
\usepackage{xcolor}          
\usepackage{multirow}
\usepackage{tablefootnote}

\usepackage{enumitem}

\usepackage{hyperref}

\algrenewcommand\algorithmicrequire{\textbf{Input:}}   
\algrenewcommand\algorithmicensure{\textbf{Output:}}   

\title{MUG-V 10B: High-efficiency Training Pipeline for Large Video Generation Models}

\author{%
Yongshun Zhang\textsuperscript{*}, \
  Zhongyi Fan\textsuperscript{*}, \
  Yonghang Zhang, \
  Zhangzikang Li, \
  Weifeng Chen,  \\
  \textbf{Zhongwei Feng,} \
  \textbf{Chaoyue Wang\textsuperscript{\textdagger}}, \
  \textbf{Peng Hou\textsuperscript{\textdagger}}, \ 
  \textbf{Anxiang Zeng\textsuperscript{\textdagger}} \\
  LLM Team, Shopee Pte. Ltd. \\
  \texttt{\{daniel.wang, peng.hou\}@shopee.com} \quad
  \texttt{zeng0118@e.ntu.edu.sg}
  \\
  Open-source repository: \url{https://github.com/Shopee-MUG/MUG-V}
}

\begin{document}

\maketitle
\begin{abstract}
  In recent years, large-scale generative models for visual content (\textit{e.g.,} images, videos, and 3D objects/scenes) have made remarkable progress. However, training large-scale video generation models remains particularly challenging and resource-intensive due to cross-modal text-video alignment, the long sequences involved, and the complex spatiotemporal dependencies. To address these challenges, we present a training framework that optimizes four pillars: (i) data processing, (ii) model architecture, (iii) training strategy, and (iv) infrastructure for large-scale video generation models. These optimizations delivered significant efficiency gains and performance improvements across all stages of data preprocessing, video compression, parameter scaling, curriculum-based pretraining, and alignment-focused post-training. Our resulting model, MUG-V 10B, matches recent state-of-the-art video generators overall and, on e-commerce-oriented video generation tasks, surpasses leading open-source baselines in human evaluations. More importantly, we open-source the complete stack, including model weights, Megatron-Core-based large-scale training code, and inference pipelines for video generation and enhancement. To our knowledge, this is the first public release of large-scale video generation training code that exploits Megatron-Core to achieve high training efficiency and near-linear multi-node scaling, details are available in \href{https://github.com/Shopee-MUG/MUG-V}{\bf our webpage}.
\end{abstract}

\begingroup
\renewcommand\thefootnote{}
\footnotetext{\textsuperscript{*} Equal contribution.}
\footnotetext{\textsuperscript{\textdagger} Corresponding author.}
\footnotetext{The rest of the authors' email address: \texttt{\{first\_name\}.\{last\_name\}@shopee.com}.}
\endgroup

\section{Introduction}
\label{sec:intro}

Generative artificial intelligence models have advanced rapidly in recent years. Scaling laws have been empirically validated in Transformer-based foundation models, yielding strong performance across multiple modalities. This rapid progress is driving a broad expansion of AIGC applications that are transforming production pipelines and daily practice, for instance, the CompassLLM series offers strong multilingual support and targeted capabilities for e-commerce.~\cite{maria2024compasslargemultilinguallanguage,maria2025compassv2technicalreport,maria2025compassv3scalingdomainspecificllms}.


In this work, we focus on developing a high-efficiency training framework for diffusion transformers (DiT) and training a large-scale video generation model. Video generation is among the most challenging forms of visual content synthesis: relative to image generation, it must preserve static content fidelity while learning diverse motion dynamics; relative to 3D generation, it should not only implicitly capture object-level 3D structure but also model inter-object interactions and physical regularities~\cite{wang2025survey,lu2024handrefiner,wiedemer2025video}. Meanwhile, training large-scale video generation models also demands addressing three core challenges: cross-modal text-video alignment, extra long visual token sequences and complex spatiotemporal patterns~\cite{wen2025analysis,zeng2024dawn,kaddour2023challenges,zhang2024vast,chen2025temporal}. 

To address these challenges, we adopt the prevailing video generation paradigm, \emph{i.e.,} latent diffusion transformers with flow matching objectives~\cite{lipman2022flow,albergo2022building}, and systematically design and implement an end-to-end training framework spanning data processing, video compression, model pre-training, post-training, infrastructure, and application evaluation. Along this pipeline, we investigate (i) how to execute each stage with high efficiency and minimal resource cost, and (ii) how to validate recent techniques and introduce techniques that further improve generative quality. Our contributions are summarized as follows:
\begin{itemize}[leftmargin=1.6em, labelsep=0.8em, itemsep=0.2ex, topsep=0.4ex]
    \item \textbf{Scalable data processing pipeline.} We built a pipeline that filters and extracts high-quality video clips from large corpora and uses a fine-tuned vision-language model (VLM) to generate structured, high-quality captions for all clips, with emphasis on throughput and stage-wise accuracy.
    \item \textbf{High-ratio VideoVAE compression.} We trained a VideoVAE that achieves $8\times8\times8$ compression along time, height, and width. Combined with non-overlapping $2\times2$ patchification in the DiT, this yields $\approx 2048\times$ compression relative to pixel space. With targeted architecture and loss design, reconstruction quality remains comparable to state-of-the-art VAEs at this ratio.
    \item \textbf{Training-stable Transformer backbone.} We designed a 10-billion-parameter Diffusion Transformer (DiT) with a transformer-block configuration that trains stably, and introduced a new image/frame conditioning scheme that improves cross-frame consistency.
    \item \textbf{Multi-stage training strategy for better video generation.} The process comprises (i) small-model hyperparameter validation, (ii) curriculum-based pre-training after scaling parameters, (iii) annealed SFT with curated high-quality data, and (iv) preference optimization using human-labeled annotations, which together substantially reduce trial-and-error compute while steadily improving performance.
    \item \textbf{Efficient training infrastructure.} Built on Megatron-Core~\cite{megatron-lm}, our system combines data, tensor, and pipeline parallelism to fully utilize hardware's compute and interconnect, avoid activation recomputation, and incorporates hand-written Triton kernels. On the system with 500 Nvidia H100 GPUs, it achieves near-linear scaling.
    \item \textbf{Full-stack open-sourcing.} We open-source the entire stack, including model weights, Megatron-Core-based large-scale training code, and inference pipelines for video generation and enhancement in \href{https://github.com/Shopee-MUG/MUG-V}{webpage}. To our knowledge, this is the first public release of large-scale video generation training code that leverages Megatron-Core for high training efficiency (\textit{e.g.}, high GPU utilization, strong MFU) and near-linear multi-node scaling. By releasing the full framework, we aim to accelerate progress in video generation and lower the barrier for researchers and practitioners to explore scalable modeling of the visual world.
\end{itemize}


\section{Data Processing}
\label{sec:data}

Compared with motion-conditioned or image-to-video supervision, video-text pairs are the primary training corpus for large video generation models: they are cheaper to collect at scale and jointly encode both visual dynamics and their semantic descriptions~\cite{chen2024panda,ju2024miradata,tan2024vidgen,wang2025koala}. 
In this work, we first built a scalable video processing pipeline that filters and captions raw footage to yield a large and diverse video clip-caption pairs. A relatively small, high-quality subset is further selected for post-training.

\subsection{Scalable Video Processing Pipeline}
\label{subsec:f_video_process}

We aggregate raw videos from both public and internal sources. Each video first undergoes a video-level screening for licensing, privacy compliance, prohibited content, and diversity of scenes, subjects, and motion. Only videos passing this gate enter the fine-grained pipeline below.

\textbf{Video splitting.} Accurately isolating semantically coherent segments is critical because current captioners struggle with clips that contain multiple scene transitions. We employ PySceneDetect~\cite{castellano2020pyscenedetect} and Color-Struct SVM (CSS) method from~\cite{wang2025koala} in tandem: PySceneDetect handles most cuts, while CSS complements it on identifying scene transitions such as gradual fades. Confidence thresholds are tuned according to data sources to maximise true-positive splits.

\textbf{Visual-quality filtering.} To guarantee sharp, aesthetically pleasing, and temporally coherent clips, we apply four-stage filtering:
\begin{itemize}[leftmargin=1.6em, labelsep=0.8em, itemsep=0.2ex, topsep=0.4ex]
    \item \textit{Sharpness test.} The Laplacian-variance metric from OpenCV~\cite{maliki2020open} highlights edge energy, frames with a variance $\in [200, 2000]$ are retained, otherwise the entire clip is rejected. 
    \item \textit{Aesthetic score.} A LAION-style aesthetic predictor discards clips scoring $< 4.5$~\cite{schuhmann2022laion,schuhmann2021laion}.
    \item \textit{Motion amplitude.} Optical flow magnitude is estimated with RAFT~\cite{teed2020raft}. We sample three evenly spaced frame pairs (start, middle, end), average their flow magnitudes, and drop clips that are nearly static ($< 1$) or overly dynamic ($> 20$).
    \item \textit{Multimodal LLM filter.} A proprietary model fine-tuned on 24k labelled videos is employed to identify clips with heavy post-processing (text overlays, large borders, special-effects), speed-altered footage, and camera shake.
\end{itemize}



\textbf{Caption generation.} High-fidelity captions are crucial for both prompt adherence and stable training convergence, which currently rely heavily on advanced video understanding models~\cite{Qwen-VL,wu2025number}. We first finetune a Qwen2-VL-72B~\cite{Qwen2VL} captioner on public datasets plus internally labelled clips, optimising for descriptions that cover objects, appearance, motion, and background context. The capability is then distilled into a Qwen2-VL-7B model~\cite{Qwen2VL}, striking a balance between accuracy and inference throughput for large-scale captioning.

\textbf{Data balancing and deduplication.} To control distributional bias and eliminate duplicates, we parse captions with a large language model that extracts key entities (subjects, actions, scenes). These tags form a lightweight ontology used to (i) stratify sampling so under-represented categories receive adequate weight and (ii) identify near-duplicate clips for removal. 

\subsection{Human-Labelled Post-training Data}
Pre-training equips the model with basic text-video alignment, motion priors, and a grasp of physical dynamics. Nevertheless, two problems persist: (i) Limited generation quality, \emph{e.g.}, low aesthetics, motion discontinuities; (ii) Physical errors, \emph{e.g.}, impossible trajectories, inconsistent details~\cite{lin2025exploring,qian2025rdpo,seawead2025seaweed}. To address these issues we curate a human-verified post-training corpus that serves two complementary purposes: (i) refining the model on the highest-quality real videos and (ii) labeling preference signals that directly target remaining failure cases.

\subsubsection{High-quality Clip Labeling}

\textbf{Score-based filtering.} From the full pre-training set we retain the top $\approx 10 \%$ of clips ranked by the composite score described in Section~\ref{subsec:f_video_process}.

\textbf{Distribution re-balancing.} Unlike the pre-training stage, we intentionally up-weight human-centric clips (people, complex body motion, human-object interactions). Our empirical finding is that rigid-object dynamics are comparatively easy to learn, whereas articulated human motion remains a major bottleneck yet dominates real user queries.

\textbf{Manual quality labeling.} Automated filters still meet failure modes such as subtle scene splices (scene transition problem) or mild video compression artefacts. Human annotators therefore review each candidate clip on three axes: (i) Motion continuity (no jump cuts or speed ramps); (ii) Content stability (no scene changes, dissolves, or stitched footage); (iii) Visual fidelity (clarity, absence of heavy post-processing). Clips failing any criterion are discarded. The resulting subset forms the supervised post-training data, offering uniformly high visual and temporal consistency.

\subsubsection{Preference Optimisation Data Labeling}
Even with real-video data training, the performance of generative model can plateau before generating reasonable videos. We thus collect human preference annotations on the model’s own outputs:

\textbf{Pairwise comparison labeling.} Annotators compare two generated videos for overall aesthetics, motion smoothness, and severity of visual errors. The preferred video receives a positive label; the other receives a negative one.

\textbf{Absolute correctness labeling.} Independently, each clip is checked for (i) semantic match to the prompt, (ii) preservation of the main subject throughout the sequence, and (iii) presence of any physical or rendering errors. These evaluations yield binary pass or fail labels.

This dual annotation scheme powers the preference-learning stage (detailed in Section~\ref{ssubsec:post_training}), enabling iterative improvement of generation quality and systematic reduction of physical errors.

\section{Model Design and Architecture}

Building on mainstream latent-diffusion and latent-flow transformer frameworks~\cite{peebles2023scalable,esser2024scaling,polyak2024movie,peng2025open,ho2022video,wan2025wan}, we adopt a two-stage generative pipeline. First, a variational auto-encoder (VAE) compresses pixel-space video frames into a compact latent representation. Next, a 10-billion-parameter Diffusion Transformer (DiT) is trained to operate entirely in this latent domain, modeling spatiotemporal dynamics to synthesize videos. To unify text-to-video and image-to-video tasks within a single architecture, we devise an image-conditioning strategy that injects visual tokens from a reference image into the context stream of DiT, allowing controllable generation conditioned on either textual prompts or key frames.

\subsection{Video Variational Autoencoder (Video VAE)}
\label{sec:video_vae}

High-quality latent representations are pivotal for training video generation diffusion models. Our Video VAE balances three objectives: (i) maximal spatiotemporal compression, (ii) preservation of fine detail, and (iii) a lightweight encoding strategy that supports rapid iteration.  
The encoder downsamples each input clip by a factor of $8 \times 8 \times 8$ along the temporal, height, and width axes, achieving a 512× volumetric compression.  
Before entering the Diffusion Transformer (DiT), we apply a non-overlapping $2 \times 2$ spatial patchification that maps every four latents to a single token.  

\subsubsection{Video VAE Architecture}
\label{subsec:vae_arch}

We initialise the Video VAE from a publicly available image VAE with strong reconstruction fidelity~\cite{podell2023sdxl} and extend it to the video domain via hybrid convolutional stacks. 
Each down-sampling stage alternates a 2D spatial convolution, capturing intra-frame texture, with a 3D convolution that models inter-frame motion. This hybrid design retains the expressiveness of a fully 3D encoder while reducing FLOPs significantly relative to an all-3D counterpart.

Unlike prior work that separates `spatial' and `temporal' pathways \cite{zheng2024open}, we adopt a unified architecture that jointly downsamples every dimension by eight. The resulting latent tensor $Z \in \mathbb{R}^{(T/8) \times (H/8) \times (W/8) \times C}$ encodes both appearance and motion cues in a compact form. Aggressive compression can harm fidelity, so we widen the bottleneck’s channel dimension to enhance latent capacity. Ablation studies show that increasing $C$ markedly improves reconstruction until diminishing returns set in, we ultimately select $C = 24$ as the best trade-off between quality and storage budget. Similar observations are reported in~\cite{seawead2025seaweed}. 

\subsubsection{Minimal Encoding Strategy}
\label{subsec:vae_minimal}

Temporal causal convolutions have become the de-facto choice in many existing Video VAE implementations because they (i) respect the arrow of time, (ii) allow a single model to encode variable-length clips, including degenerate cases such as still images or first-frame conditioning, and (iii) prevent information `leakage' from future frames during video prediction. However, causal convolutions also introduce some drawbacks. When the distance from the current frame to the clip origin is smaller than the encoder’s temporal receptive field, earlier tokens aggregate less context than later ones, yielding an information imbalance across the latent sequence.  
Meanwhile, for clips whose length differs from the receptive field, the imbalance persists even after remedies such as fixed-length windows with overlapping weighted sums.

\textbf{Minimal-Encoding Principle.} To eliminate these issues, we proposed the \emph{minimal encoding principle} for Video VAE. Specifically, we enforce that each latent token as an independent unit derived solely from its corresponding frame chunk (8 in our setting), thus no information is exchanged beyond this temporal window. We argue that the primary responsibility of Video VAE are compression and reconstruction, yet not generation. Thus, because the unit frame segment already contains the appearance and motion cues required to reconstruct itself, further context mixing is unnecessary and may even create shortcut learning. The minimal principle also yields a flexible latent interface: the same encoder can be used for arbitrary sequence lengths, for image‐to-video or video continuation tasks, and for special cases such as first-, middle-, or last-frame conditioning.

\textbf{Sharing Decoder Strategy.}  The decoder must reconstruct the complete clip from the latent sequence, it is not bound by the above `minimal principle'. Empirically, feeding an appropriate span of latents to the decoder in one shot leads to faster convergence than forcing unit-wise reconstruction.  
To balance throughput and memory, we train with single-latent encoding but vary the decoder’s input window across $\{1, 4, 8\}$ contiguous latents. At run time the encoder and decoder simply reshape their inputs to match the chosen window size (see Appendix~\ref{append:videovae} Algorithm~\ref{alg:minimal_vae}).

This minimal-encoding design removes the information-density imbalance of causal convolutions while retaining compatibility with downstream tasks and diverse clip lengths, contributing significantly to MUG-V 10B’s overall training efficiency.

\subsection{MUG-V 10B Diffusion Transformer Model}

The generative core of MUG-V is a 10-billion-parameter Diffusion Transformer. The model is trained jointly for text-to-video, image-to-video, and text-plus-image-to-video synthesis, thereby unifying the principal conditioning modalities required for modern video generation. Its backbone follows the DiT architecture~\cite{peebles2023scalable}, ensuring compatibility with state-of-the-art diffusion techniques. Our DiT backbone consists of four components: (i) input patchifying, (ii) text condition networks, (iii) stacked DiT blocks, and (iv) output unpatchifying. Its overall organisation follows some existing DiT models~\cite{peebles2023scalable,polyak2024movie,zheng2024open}, so this report focuses on specific design choices rather than restating the entire architecture.

\textbf{Transformer block.} Instead of the MM-DiT block used in some image/video diffusion models~\cite{esser2024scaling,kong2024hunyuanvideo}, we adopt a transformer block architecture closely aligned with that of autoregressive language models. A cross-attention module is inserted between the self-attention and feed-forward network (FFN) to enable direct interaction between textual embeddings and visual tokens.

\textbf{Full attention v.s. spatio-temporal separated attention.} Current DiT variants employ either full attention~\cite{wan2025wan}, where every token in the spatiotemporal sequence attends to every other, or spatio-temporal separated attention~\cite{gao2025seedance}, which restricts attention to a local neighbourhood to reduce computation. Full attention provides stronger global coherence, for example, the same person or background appearing at the start and end of a clip can interact directly. Because our Video VAE and patchifying scheme yield a high compression ratio, full attention does not incur prohibitive cost, so we adopt it throughout.

\textbf{3D RoPE encoding for visual tokens.} To allow full attention to capture accurate positional cues, we apply three-dimensional Rotary Position Embedding (RoPE), which extends the original 1D formulation to jointly encode spatial and temporal coordinates~\cite{su2024roformer}.

\textbf{Global signal embedding.} Global signals such as diffusion timesteps and video frame-rate are embedded following~\cite{esser2024scaling}. A shared MLP maps each global scalar to the model dimension, and per-block learnable scale parameters modulate the resulting vector, balancing expressiveness with memory efficiency.

\textbf{Normalisations.} Consistent with prior large-scale models, normalisation improves training stability. Beyond the QK normalisation inside self-attention, we normalise input text features and the cross-attention module~\cite{ba2016layer,zhang2019root}. Empirically, these layers markedly reduce parameter volatility and attenuate loss fluctuations, leading to fewer visual artefacts during the training procedure.

\textbf{Image/frame conditioning.} For image- or frame-conditioned video generation, we mask the video sequence rather than add conditional latents to the denoising latent. Conditioned regions receive the given image/frame latent and have their diffusion timestep set to zero (zero noise added), while the remaining tokens follow the standard noisy diffusion trajectory. During pre-training this strategy both clarifies the timestep signal and yields superior fidelity to the provided visual content at inference.

\section{Model Training}

\subsection{Video VAE Training}
\label{subsec:vae_loss}

The Video VAE is trained with the composite loss,
\begin{equation}
  \mathcal{L}_{\text{VAE}} = 
  \mathcal{L}_{\text{rec}} + 
  \lambda\,\mathcal{L}_{\text{KL}} +
  \gamma\,\mathcal{L}_{\text{GAN}},
\end{equation}
where the three terms serve complementary purposes:

\begin{itemize}[leftmargin=1.6em, labelsep=0.8em, itemsep=0.2ex, topsep=0.4ex]
  \item \textbf{Reconstruction loss}  
        $\mathcal{L}_{\text{rec}}$ is a weighted sum of 
        $\mathcal{L}_\text{MSE}$, $\mathcal{L}_1$, and $\mathcal{L}_{\text{perc}}$, we  
        encourage pixel-level accuracy (MSE, $\ell_1$) and perceptual fidelity ($\mathcal{L}_{\text{perc}}$).

  \item \textbf{Kullback-Leibler divergence}  
        $\mathcal{L}_{\text{KL}}$ regularises the latent distribution, suppressing outliers and promoting smooth interpolation.

  \item \textbf{Adversarial loss}  
        $\mathcal{L}_{\text{GAN}}$ is applied only during the final fine-tuning stage to sharpen texture and colour.  
        Because excessive adversarial weighting can introduce hue shifts or over-enhanced details, we keep $\gamma$ small and monitor validation PSNR/SSIM.
\end{itemize}

\paragraph{Adaptive Reconstruction Weighting.}
After the core objective stabilises, we observe that the model readily reconstructs global structure but oscillates on highly dynamic, fine-detail regions.  
To focus learning on these harder cases, we introduce an \emph{adaptive reconstruction loss}.  

For each reconstructed frame $x_t$ we compute a spatiotemporal saliency map
\[
w_t \;=\; \bigl|
           \,\, \Delta_t \!\bigl(\,\nabla^2 x_t\bigr)
           \,\,\bigr|,
\]
where $\nabla^2$ is the Laplacian (extracting high-frequency spatial edges) and $\Delta_t$ is the temporal forward difference (highlighting fast motion). Then, we employ $w_t$ to form the weighted loss term $\mathcal{L}_{\text{adaptive}}$ to replace the plain $\ell_1$ component in $\mathcal{L}_{\text{rec}}$.  
Regions with rapid spatiotemporal change thus contribute a larger gradient signal, improving convergence without additional data passes.


\subsection{MUG-V 10B Diffusion Transformer Training}
\label{sec:Dit_training}

To achieve high training efficiency and maintain convergence stability at this scale, besides the model architectural refinements, we incorporate three technical measures: (i) a parameter-expansion strategy accompanied with systematic hyper-parameter search; (ii) a multi-stage pre-training curriculum; and (iii) supervised and preference optimization based post-training. These designs enable stable and resource-efficient training of the 10B parameter DiT without compromising video generation quality.

\subsubsection{Parameter Expansion}\label{sec:parameter-expansion}

Considering that perform exhaustive scaling law studies and hyper-parameter sweeps would cost plenty of computing resources, we adopted a two-stage workflow: first train a compact model, then expand its parameters to the 10B scale for continued training.

Similar to zero-shot hyper-parameter transfer researches~\cite{yang2021tuning,zheng2025scaling}, we fixed the target depth at 56 Transformer blocks and built a smaller DiT with hidden size~1728 (leading to a approximate 2B parameters model). Its low training cost and fast inference made it ideal for rapid experimentation and recipe validation. Once this 2B model achieved satisfactory video-generation quality, we enlarged it via a hidden-size equi-variant expansion.

Our strategy closely related to the \emph{HyperCloning} expansion method~\cite{samragh2024scaling}, we both increase channel width while preserving the network’s functional behaviour. Consider a linear layer with weights \(W\in\mathbb{R}^{d\times d}\) and bias \(b\in\mathbb{R}^{d}\). Expanding the hidden dimension by a factor \(e\) produces \(W'\in\mathbb{R}^{ed\times ed}\) and \(b'\in\mathbb{R}^{ed}\) by tiling the original parameters and dividing by \(e\) to keep feature scaling unchanged. Meanwhile, random perturbations are added to avoid the gradient duplication problem. Thus,
\begin{equation}
W'=\frac{1}{e} (
\begin{pmatrix}
W & \cdots & W\\
\vdots & \ddots & \vdots\\
W & \cdots & W
\end{pmatrix} -
\begin{pmatrix}
\epsilon_{11} & \cdots & \epsilon_{1n}\\
\vdots & \ddots & \vdots\\
\epsilon_{m1} & \cdots & \epsilon_{nm}
\end{pmatrix}),
\qquad
b'=\frac{1}{e}
\begin{pmatrix}
b\\ \vdots\\ b
\end{pmatrix}.
\end{equation}
Inputs are replicated,
\(x_i'=[\,x_i; \ldots; x_i\,]\), and the outputs
\(x_o' \approx [\,x_o; \ldots; x_o\,]\),
each repeated \(e\) times. Setting \(e=2\) increased the total parameter count by roughly \(4\times\). After initialising the 10B model with these expanded weights, we transferred the hyper-parameters tuned on the 2B model and resumed training~\cite{yang2021tuning,mccandlish2018empirical,zheng2025scaling}. This output-preserving expansion accelerated convergence, while the small-model stage substantially reduced overall experimentation cost.

\subsubsection{Multi-stage Pre-training Curriculum}\label{sec:curriculum}

The heterogeneous nature of video data, where low-level texture and high-level semantics coexist, makes curriculum learning particularly effective for training video generation model. At low spatial resolution, semantic content dominates, as resolution increases, richer textures emerge. Moreover, a video can be viewed as a dynamic extension of static image, with motion learned on top of appearance. Leveraging these properties, we adopt a three-stage curriculum: 
\begin{itemize}[leftmargin=1.6em, labelsep=0.8em, itemsep=0.2ex, topsep=0.4ex]
    \item {Stage 1} mixes image data with low-resolution (360p) video clips. The image-to-video ratio is annealed during training until video dominates, at which point the model reliably produces plausible images and coarse video clips.  
    \item {Stage 2} retains the 360p resolution but increases clip length from 2s to 5s, and training continues until the validation loss plateaus.  
    \item {Stage 3} replaces the training set with 5s clips at 720p, curated from around 12M high-quality videos, constituting the final pre-training phase.  
\end{itemize}

Note that (i) the relatively small model before parameter-expansion use only images and 360p videos; (ii) aforementioned masking strategy for image/frame conditioning is compatible with the text-to-video generation pretraining, and we introduce the first frame masking in both stage 2 and 3.

This curriculum not only guides the model to acquire video-generation skills progressively but also boosts training efficiency. In Stages 1 and 2, shorter sequences and higher throughput allow the model to see over ten times more samples than in Stage 3, fostering robust general abilities. Stage 3, although computationally costly, refines detail thanks to its rigorously filtered, high-resolution data.

\subsubsection{Post-training and Alignment}
\label{ssubsec:post_training}

After the multi-stage pre-training, the validation loss plateaued and began to oscillate, the model’s outputs exhibited two persistent failure modes: (i) fine-grained artefacts, especially in articulated regions such as human hands, (ii) violations of basic physical plausibility (\emph{e.g.,} interpenetration and distortions). To further improve generative quality we adopted two post-training approaches: annealed supervised fine-tuning (SFT) with post-EMA, and preference-based optimization~\cite{karras2024analyzing,rafailov2023direct,peng2025lmm}.

\paragraph{Annealed SFT with post-EMA.}
We first refined the training corpus, manually selecting around 0.3 M high-quality clips. Continuing training on this subset with a gradually decaying learning rate proved effective. We compared online exponential-moving-average (EMA) parameter smoothing with a post-hoc EMA~\cite{karras2024analyzing} variant. The latter not only removed the need for expensive grid search over EMA hyper-parameters but also more likely to produce higher video quality. Instead of the post-hoc EMA proposed by Karras \textit{et al.}~\cite{karras2024analyzing}, we approximate it by exponentially decayed model ensembling, which is conceptually similar to the \textit{model merging} strategy in~\cite{li2025model} and empirically outperforms standard online EMA in our setting.

\paragraph{Preference optimisation.}
Although preference-based reinforcement learning has achieved notable success in large language models, its application to video generation remains challenging due to (i) the limited capacity of current video evaluation (reward) models and (ii) the multiplicity of optimization axes, such as appearance, motion, temporal coherence, and so on. We therefore resorted to human-annotated preferences, focusing on two objectives:
\begin{itemize}[leftmargin=1.6em, labelsep=0.8em, itemsep=0.2ex, topsep=0.4ex]
  \item \textit{Error-free generation.} For failures such as interpenetration, deformation, or other physical implausibilities we collected absolute positive/negative labels and optimised the model with the KTO algorithm~\cite{ethayarajh2024kto,li2024aligning}.
  \item \textit{Motion quality.} To improve dynamic realism we obtained pairwise ``better/worse'' annotations and applied the DPO algorithm~\cite{rafailov2023direct,wallace2024diffusion}.
\end{itemize}
Retaining the original supervised fine-tuning (SFT) objective as a regularizer during preference optimization mitigated the risk of the model adopting undesirable statistical biases (\emph{e.g.,} exaggerated motion amplitude or recurring texture patterns). Conducting preference optimization in multiple stages and interleaving batches from different annotation sources allowed the model to sequentially expose distinct classes of errors, thereby achieving continuous quality improvements.

\section{Infrastructure}

Beyond algorithmic design, infrastructure is pivotal to achieving efficient and stable training for large-scale video generation. Our video generation DiT model faces \emph{processing long sequences with full attention}, \emph{scaling to billions of parameters}, and \emph{preserving numerical precision during training} three core challenges. We therefore build a Megatron-Core~\cite{megatron-lm} based training framework for MUG-V 10B, concentrating on three optimizations: (i) model-parallel strategy, (ii) balanced data-loading/training pipelines, and (iii) fused kernel, to overcome these obstacles.


\subsection{Model Parallel Strategy}

Given the long-sequence nature of video data, which incurs higher dynamic memory consumption than language models' pretraining, we systematically explored parallelization techniques to maximize throughput. Our hybrid scheme combines data parallelism (DP), tensor parallelism (TP), pipeline parallelism (PP), and sequence parallelism (SP).

To train our 10B DiT model, we first enable TP within a single node. To alleviate the memory burden of long sequences, we shard activations across the TP group via SP. Next, we apply PP, vertically partitioning layers and leveraging point-to-point communication to exploit inter-node bandwidth while disabling activation recomputation. Finally, we introduce DP to enlarge the effective batch size and improve training stability. Extensive benchmarking identifies an optimal 10B-scale configuration that delivers near-linear efficiency scaling, thereby maximizing hardware utilization.

\subsection{Data Loading and Computation Balance}

Beyond optimizing parameter updates, efficient data ingestion is crucial to overall training throughput. We build an asynchronous I/O pipeline with aggressive pre-fetching and caching, overlapping data preprocessing and transfer with computation to hide latency. To minimize pipeline stalls arising from variable video sequence lengths, we also introduce dynamic balanced sampling across all ranks. This scheme ensures that each GPU receives batches of comparable computational cost, reducing idle cycles and further improving hardware utilization.

\subsection{Kernel Fusion}

To reduce DiT’s memory overhead from pixel-wise modulation and residual paths, we design a two-tier fusion of low-level kernels and block refactoring.

We merge three tightly coupled operations, (i) linear-layer bias addition, (ii) per-pixel scale-and-shift modulation, and (iii) residual accumulation into a single GPU kernel. Collapsing the read-compute-write sequence into one pass cuts global-memory transactions from N down to one. The fused kernel is handwritten in Triton, leveraging warp-level shuffles to broadcast bias and modulation vectors without shared-memory spills. A persistent-threads scheduling pattern keeps intermediate data resident in registers across the three fused stages, pushing bandwidth utilisation toward hardware limits and further trimming memory traffic.

At a higher level, we reshape the DiT block to expose additional fusion opportunities:
\begin{itemize}[leftmargin=1.6em, labelsep=0.8em, itemsep=0.2ex, topsep=0.4ex]
    \item LayerNorm + QKV Projection. Layer normalization is executed in tandem with the query-key-value (QKV) projection, eliminating an extra memory round-trip.
    \item Masked Softmax Fusion. Attention-score masking is folded directly into a FlashAttention-2 soft-max kernel, avoiding redundant reads of the score matrix.
    \item Zero-Padding Removal. Static shape inference removes unnecessary padding, ensuring fully coalesced accesses.
\end{itemize}
Together, these optimizations lower memory traffic, increase arithmetic intensity, and deliver an end-to-end speed-up.

\section{Applications \& Model Performance}
\label{sec:exp}

Video-generation technology is now routinely applied in film, gaming, advertising, and e-commerce, where it offers substantial gains in creativity and cost efficiency. As an e-commerce company, we focus on retail-specific situations: generating dynamic product videos such as try-on showcases, still-life displays, functional demonstrations, and advertising assets. To be viable in this setting, generated videos must exhibit (i) generative correctness (semantically accurate content and physically plausible motion), (ii) content consistency, and (iii) visual appeal. These requirements largely align with established evaluation protocols, so we first benchmark our models with standard automatic metrics. However, we find that existing metrics often overlook fine-grained defects, \emph{e.g.,} altered fabric textures or incorrect hand poses, that are critical for product fidelity. We therefore supplement automatic scores with human evaluations to judge overall usability and quality.

\begin{table}[t!] \vspace{-0.5 cm}
 \caption{Quantitative comparisons of video genertion\tablefootnote{Given that VBench is a widely used benchmark for video-generation evaluation, we submitted our results to enable direct comparison with prior methods. We present a subset of the VBench-I2V leaderboard, restricted to recent methods accompanied by a technical report. The complete leaderboard is available at \href{https://huggingface.co/spaces/Vchitect/VBench_Leaderboard}{this link}.}.}
  \centering
  {\scriptsize
  \begingroup
  \setlength{\tabcolsep}{4pt}
  \begin{tabular}{c|c|ccc|cccccc|cc|c}
    \toprule
        \multirow{2}{*}{Model} & {Model} & \multirow{2}{*}{VTCM} & \multirow{2}{*}{VISC} & \multirow{2}{*}{VIBC} & \multirow{2}{*}{SC} & \multirow{2}{*}{BC} & \multirow{2}{*}{MS} & \multirow{2}{*}{DD} & \multirow{2}{*}{AQ} & \multirow{2}{*}{IQ} & {I2V} & {Quality} & {Total} \\
         & {Size} & & & & & & & & & & {Score} & {Score} & {Score} \\
    \midrule
    CogVideoX~\cite{yang2024cogvideox} & 5b & \underline{67.68} & 97.19 & 96.74 & 94.34 & 96.42 & 98.40 & 33.17 & 61.87 & 70.01 & 94.79 & 78.61& 86.70\\
    STIV~\cite{lin2024stiv} & 8.7b & 11.17 & \textbf{98.96} & 97.35 & \textbf{98.40} & \underline{98.39} & \textbf{99.61} & 15.28 & \textbf{66.00} & \textbf{70.81} & 93.48 & 79.98 & 86.73\\
    Step-Video~\cite{huang2025step} & 30b & 49.23 & 97.86 & 98.63 & \textit{96.02} & 97.06 & \underline{99.24} & 48.78 & 62.29 & \underline{70.44} & \textit{95.50} & \textit{81.22} & 88.36 \\
    Dynamic-I2V~\cite{liu2025dynamic} & 5b & \textbf{88.10} & \underline{98.83} & \textit{98.97} & \underline{96.21} & \underline{98.39} & 98.88 & 27.15 & 60.10 & 69.23 & \textbf{98.12} & 78.78 & \emph{88.45} \\
    HunyuanVideo~\cite{kong2024hunyuanvideo} & 13b & 49.91 & 98.53 & 97.37 & 95.26 & 96.70 & \textit{99.23} & 22.20 & 62.55 & \textit{70.14} & 95.10 & 78.54 & 86.82 \\
    Wan2.1~\cite{wan2025wan} & 14b & 34.76 & 96.95 & 96.44 & 94.86 & \textit{97.07} & 97.90 & \textit{51.38} & \underline{64.75} & \underline{70.44} & 92.90 & 80.82 & 86.86 \\
    MAGI-1~\cite{teng2025magi} & 24b & \textit{50.85} & 98.39 & \underline{99.00} & 93.96 & 96.74 & 98.68 & \textbf{68.21} & \textit{64.74} & 69.71 & \underline{96.12} & \textbf{82.44} & \textbf{89.28}\\
    MUG-V(Ours) & 10b & 23.17 & \textit{98.82} & \textbf{99.51} & 95.73 & \textbf{98.52} & 98.90 & \underline{57.24} & 61.37 & 68.48 & 95.37 & \underline{81.55} & \underline{88.46} \\
    \bottomrule
  \end{tabular} \vspace{-0.6 cm}
  \endgroup
  \label{tab:table_vbench}}
\end{table}

\subsection{Quantitative Evaluation of Video Generation}

To evaluate the quality of videos generated by MUG-V 10B, especially the text-image to video(TI2V) setting emphasized in e-commerce, we adopt the VBench protocol and related metrics. We assess overall quality along three dimensions, \textit{i.e.,} temporal consistency, motion dynamics, and perceptual aesthetics/distortion, using six metrics: Subject Consistency (SC), Background Consistency (BC), Motion Smoothness (MS), Dynamic Degree (DD), Aesthetic Quality (AQ), and Imaging Quality (IQ). Additionally, three I2V-specific metrics are included: Video-Text Camera Motion (VTCM), Video-Image Subject Consistency (VISC), and Video-Image Background Consistency (VIBC). The final VBench score is computed as a weighted sum of these components~\cite{huang2024vbench,huang2024vbench++}. In our experiments, we strictly follow the VBench-I2V evaluation and submit results to the public leaderboard. As shown in Table~\ref{tab:table_vbench}, our model performs strongly across almost all metrics. At submission time, MUG-V 10B ranks third on the VBench I2V leaderboard, behind Magi-1 and the commercial system PI.

\subsection{Human Evaluation on E-commerce Video Generation Tasks}
To more directly compare against leading open-source model, HunyuanVideo and Wan 2.1, we conducted a human evaluation tailored to e-commerce video generation. Test inputs were randomly sampled from publicly available model showroom images. For each method, we used its default prompt generator to create video prompts and produced 5 seconds clips. All clips were pooled and randomly ordered, then evaluated in parallel by three independent annotators, final labels were determined by consensus (\emph{i.e.,} $\geq 2$ of 3).

The annotation proceeded in three stages. First, annotators judged whether a clip was discernibly AI-generated, considering both the presence of errors (from physical implausibilities to minor artifacts) and overall visual realism. Second, for clips deemed sufficiently realistic, annotators assessed product consistency relative to the input image, requiring that color, material, texture, and other attributes remain unchanged. We consider a clip deployable in e-commerce only if it satisfies these two criteria. Third, for deployable clips, annotators judged whether the video is “high quality,” defined by the hallmarks of professional cinematography and model performance. Finally, our model achieves strong results on both the pass rate and the high-quality rate. Since the space limitation, the detail evaluation results are reported in Appendix~\ref{append_human_eval}. Nevertheless, we observe that residual minor artifacts and geometric distortions still limit overall quality, indicating substantial headroom for improvement in e-commerce applications.

\section{Related Works}

\subsection{Diffusion Models}

Diffusion-based generative modeling originates from score matching and denoising autoencoders, culminating in denoising diffusion probabilistic models (DDPM) and the continuous-time score-based SDE/ODE formulations~\cite{ho2020denoising,song2020score,song2020denoising}. These methods learn a reverse-time denoising process to transform Gaussian noise into data, and support conditioning through classifier/classifier-free guidance as well as latent-space diffusion with learned encoders for efficiency~\cite{dhariwal2021diffusion,rombach2022high,zhao2023null}.

A subsequent line of work replaces stochastic reverse diffusion with deterministic transport, framing generation as learning a velocity field that pushes a simple prior to the data distribution. Rectified flow and flow matching objectives directly supervise this transport via continuity equations or optimal-transport-inspired training, often yielding faster sampling and simpler training dynamics~\cite{lipman2022flow,albergo2022building}. Complementary advances distillation to few/one-step samplers, consistency models, improved solvers, and DiT backbones—further reduce inference cost while preserving fidelity~\cite{salimans2022progressive,song2023consistency,peebles2023scalable}.

Diffusion and flow matching have been successfully applied across modalities. In images, latent diffusion enabled text-conditional, high-resolution synthesis at scale~\cite{podell2023sdxl,hu2022unified,esser2024scaling,he2024diff}, DiT backbones improved scaling and training stability~\cite{peebles2023scalable}. In 3D, score-distillation and related techniques optimize neural or explicit 3D representations from text or image supervision~\cite{poole2022dreamfusion,wang2023prolificdreamer}. Audio and music generation commonly operate in spectrogram space with text or melody conditioning~\cite{kong2020diffwave,agostinelli2023musiclm}. Motion, trajectories, and robotics have leveraged diffusion priors for controllable dynamics~\cite{chi2023diffusion}. Extensions to discrete domains (code or text) use relaxed tokenizations or hybrid AR-diffusion designs~\cite{austin2021structured,li2022diffusion,song2025seed}. These developments establish diffusion/flow matching as flexible, scalable foundations for high-dimensional generative tasks, including video.

\subsection{Video Generation Models}

Early text-to-video systems extended image diffusion with temporal priors or cascaded frame synthesis and super-resolution, but were limited in duration, resolution, and temporal coherence~\cite{singer2022make,ho2022imagen,villegas2022phenaki}. Latent video diffusion improved efficiency by compressing videos with video VAEs before applying spatiotemporal denoising~\cite{xing2024large,chen2024deep}, enabling longer clips and higher fidelity~\cite{blattmann2023stable}.

A dominant family today uses DiT-based generators operating in a latent video space: a video VAE provides compact spatiotemporal latents, while a DiT (with factorized or windowed spatiotemporal attention) performs conditional generation under text, image, or control signals~\cite{zhou2022magicvideo,chen2023videocrafter1}. Advances include stronger conditioning (pose, depth, camera paths, audio), longer context handling (memory-efficient attention, sliding windows), and faster sampling via consistency or flow matching objectives. Large proprietary systems, \emph{e.g.,} Sora~\cite{sora_openai_2024}, Sora2~\cite{sora2_openai_2025}, Veo3~\cite{deepmind_veo3_models_2025} demonstrate long-duration, high-resolution generation with improved physical plausibility through large-scale training, aggressive latent compression, and optimized inference~\cite{brooks2024video,wan2025wan,kong2024hunyuanvideo,gao2025seedance,ma2025step,chen2025goku,team2025yan}. 

In parallel, autoregressive (AR) based approaches tokenize videos with vector-quantized encoders and model generation as next-token prediction across spatiotemporal tokens~\cite{yan2021videogpt,yu2023magvit}. These models integrate naturally with multimodal LLMs and show strengths in long-horizon structure and discrete controllability, but often trade off visual fidelity and suffer from compression artifacts. Hybrid systems combine AR planning (for structure and semantics) with diffusion/flow decoders (for photorealism), narrowing this gap~\cite{kondratyuk2023videopoet,chen2024videocrafter2,wang2025ovis,xie2025show,agarwal2025cosmos}.

Positioned within this landscape, MUG-V 10B follows the DiT-in-latent-video paradigm with an emphasis on efficient training (Video VAE compression, kernel/system optimizations) and modern training curriculum, while targeting practical conditioning modes (text-to-video and image-to-video) and alignment for e-commerce content.

\section{Conclusion}

In this report, we presented the training framework of MUG-V 10B diffusion transformer (DiT) model for video generation. Under constrained compute, we pursued an end-to-end design that integrates scalable data processing, a high-compression Video VAE, a DiT-based generator, multi-stage pre-training and post-training, and systems-level optimizations for efficient training and evaluation. Our study not only validates several recent advances for large-scale DiT model training, but also introduces practical strategies that stabilize optimization and improve generated sample quality. Across qualitative and quantitative evaluations, MUG-V 10B delivers competitive or superior performance, particularly in e-commerce scenarios. 



\newpage

\appendix

\section{Additional Technical Details}
In this section, we provide additional technical details that were omitted from the main text due to space constraints.

\subsection{More Details of Video VAE}
\label{append:videovae}
Algorithm~\ref{alg:minimal_vae} presents the pseudocode of the Video VAE Minimal Encoding Strategy. As shown, the proposed strategy can be implemented with a simple tensor reshape operation, introducing no additional computational overhead to the overall process.

\begin{algorithm}[h]
  \caption{Video VAE Minimal Encoding Strategy}
  \label{alg:minimal_vae}
  \begin{algorithmic}[1]
    \Require $\mathbf{V}\in\mathbb{R}^{1 \times T\times C\times H\times W}$, $R\in \{1,4,8\} $ \Comment{an input video $\mathbf{V}$, and the decoder window $R$}
    \State \textbf{$\mathbf{V}_\text{in}\gets \mathrm{reshape}(\mathbf{V},\,[T / 8,\, 8])$} \Comment{($T / 8) \times 8 \times C \times H \times W$}
    \State $\mathbf{E}\gets \mathrm{VaeEncoder}(\mathbf{V}_\text{in})$ 
    \State $(\boldsymbol{\mu},\boldsymbol{\log\sigma^{2}})\gets \mathrm{Linear}(\mathbf{E})$ 
    \State $\mathbf{z}\gets \mathrm{reparameterize}(\boldsymbol{\mu},\boldsymbol{\log\sigma^{2}})$ \Comment{$(T / 8) \times 1 \times C_z$} 
    \State \textbf{$\mathbf{D}\gets \mathrm{reshape}(\mathbf{z},\,[(T/8)/R,\,R])$} \Comment{$(T / 8) / R \times R \times C_z$}
    \State $\hat{\mathbf{V}}\gets \mathrm{VaeDecoder}(\mathbf{D})$ \Comment{$(T / 8) / R \times (8 \cdot R) \times C \times H \times W$}
    \Ensure $\mathrm{reshape}(\hat{\mathbf{V}})$  \Comment{reconstructed video, $1 \times T\times C\times H\times W$}
  \end{algorithmic}
\end{algorithm}

\begin{table}[b]
 \caption{Quantitative comparisons of video reconstruction.}
  \centering
  \begingroup
  \setlength{\tabcolsep}{5.5pt}
  \begin{tabular}{ccccccc}
    \toprule
    \multirow{2}{*}{Model} & Downsample & \multirow{2}{*}{Res.} & \multicolumn{4}{c}{Evaluation Metrics}\\
    & Factor & & PSNR($\uparrow$) & SSIM($\uparrow$) & LPIPS($\downarrow$) &  FloLPIPS($\downarrow$)           \\
    \midrule
    Opensora VAE & $4\times 8 \times 8$ & 256p & 28.2 & 0.821 & 0.114 & 0.108 \\
    CogVideoX VAE & $4\times 8 \times 8$ & 256p & 30.3
     & 0.902 & 0.055 & 0.053 \\
    MUG-V VAE & $8\times 8 \times 8$ & 256p & \textbf{32.2}
     & \textbf{0.912} & \textbf{0.053} & \textbf{0.048} \\
    \midrule
    Opensora VAE & $4\times 8 \times 8$ & 480p & 30.0 & 0.857 & 0.107 & 0.101 \\
    CogVideoX VAE & $4\times 8 \times 8$ & 480p & 30.5
     & 0.918 & 0.044 & 0.045 \\
    MUG-V VAE & $8\times 8 \times 8$ & 480p & \textbf{31.2}
     & \textbf{0.911} & \textbf{0.043} & \textbf{0.041} \\
    \midrule
    Opensora VAE & $4\times 8 \times 8$ & 720p & 30.6 & 0.866 & 0.109 & 0.105 \\
    CogVideoX VAE & $4\times 8 \times 8$ & 720p & 31.8 & \textbf{0.912} & 0.058 & 0.058 \\
    MUG-V VAE & $8\times 8 \times 8$ & 720p & \textbf{32.9}
     & 0.911 & \textbf{0.056} & \textbf{0.056} \\
    \bottomrule
  \end{tabular}
  \endgroup
  \label{tab:table_vae}
\end{table}

\subsection{More Details of Preference Optimisation}
\label{append:po}

\begin{figure}
  \centering
\includegraphics[width=\textwidth]{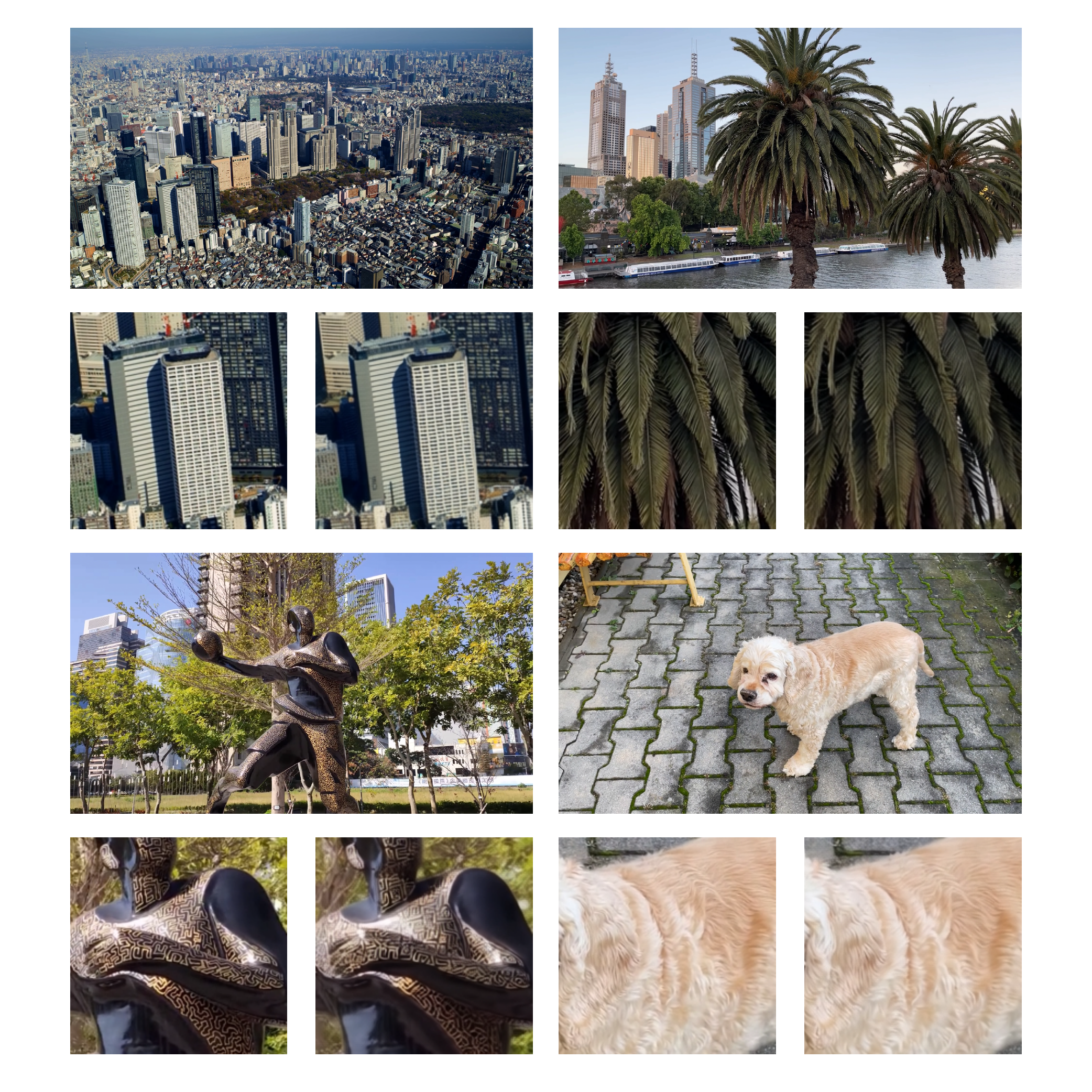} 
\vspace{-0.8cm}
    \caption{
    The visualization of Video VAE reconstruction examples. For each example, we provide the input video frame (the whole frame and local details) and the local patch extracted from the reconstructed video clip (the right enlarge part).
    }
    \label{fig:vae} 
\end{figure}

\begin{figure}
  \centering
\includegraphics[width=0.7\textwidth]{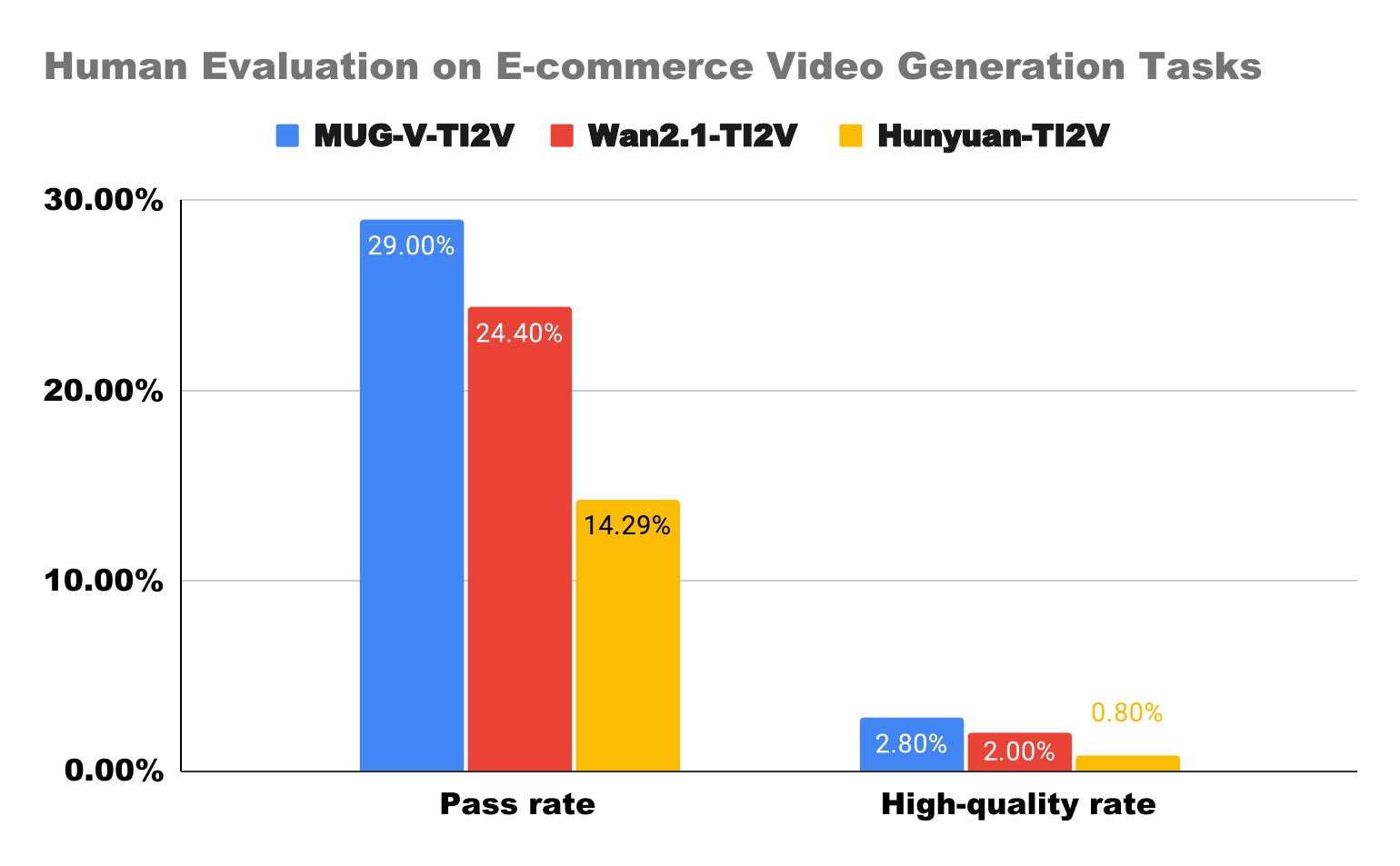} 
    \caption{
    The human evaluation comparisons on generated e-commerce video of their quality.
    }
    \label{fig:human_eval} 
\end{figure}

In addition to direct preference optimization (DPO) and KTO with human-labeled data, we introduce Real Data Preference Optimization (RDPO)~\cite{qian2025rdpo}. By applying reverse sampling on real video data, we observe that the flow sampling paths derived from these reverse samples are statistically superior to those obtained from randomly initialized noise and its associated flow trajectories. This property allows RDPO to automatically construct preference pairs without the need for manual annotation. Furthermore, a multi-stage iterative training schedule is employed to progressively improve the generator’s performance. During post-training, we observe that well-tuned reinforcement learning can rapidly strengthen the model’s generative capabilities along specific dimensions, however, striking an effective trade-off between supervised fine-tuning (SFT) and RL remains a challenging open problem~\cite{wu2025generalization}.

\section{Additional Experiments}

\subsection{Video VAE Reconstruction}

Within our video generation pipeline, the Video VAE is dedicated to compressing the video signal and reconstructing it. We therefore curated a set of real-world clips for validation and evaluated reconstruction fidelity with standard metrics, PSNR, SSIM, LPIPS, and FloLPIPS, against several baseline VAE models. As summarized in Table~\ref{tab:table_vae}, our Video VAE surpasses most comparators on these metrics. Although its score on SSIM (720p setting) is slightly lower than that of CogVideoX VAE, it delivers an $8 \times 8 \times 8$ compression ratio, achieving a favorable efficiency-quality balance. Qualitative examples in Fig.~\ref{fig:vae} show that fine details such as drifting smoke and rapidly changing textures are faithfully reproduced.

\subsection{Evaluation Results of E-commerce Video Generation}
\label{append_human_eval}
In Fig.~\ref{fig:human_eval}, we report the evaluation results of different models on e-commerce video generation, measured by pass rate and high-quality rate. In mixed blind evaluations, our MUG-V 10B achieves superior scores. Specifically, the higher pass rate indicates that our model generates a larger proportion of e-commerce videos without noticeable artifacts or errors, making them indistinguishable from real footage. Meanwhile, the improved high-quality rate reflects better performance in terms of motion coherence, visual fidelity, and aesthetics. Nonetheless, the relatively low absolute values of both metrics highlight that substantial room for improvement remains, underscoring the need for further advancement in video generation models, including ours.

\subsection{Visualizing Generated Videos}
\begin{figure}
  \centering
\includegraphics[width=0.95\textwidth]{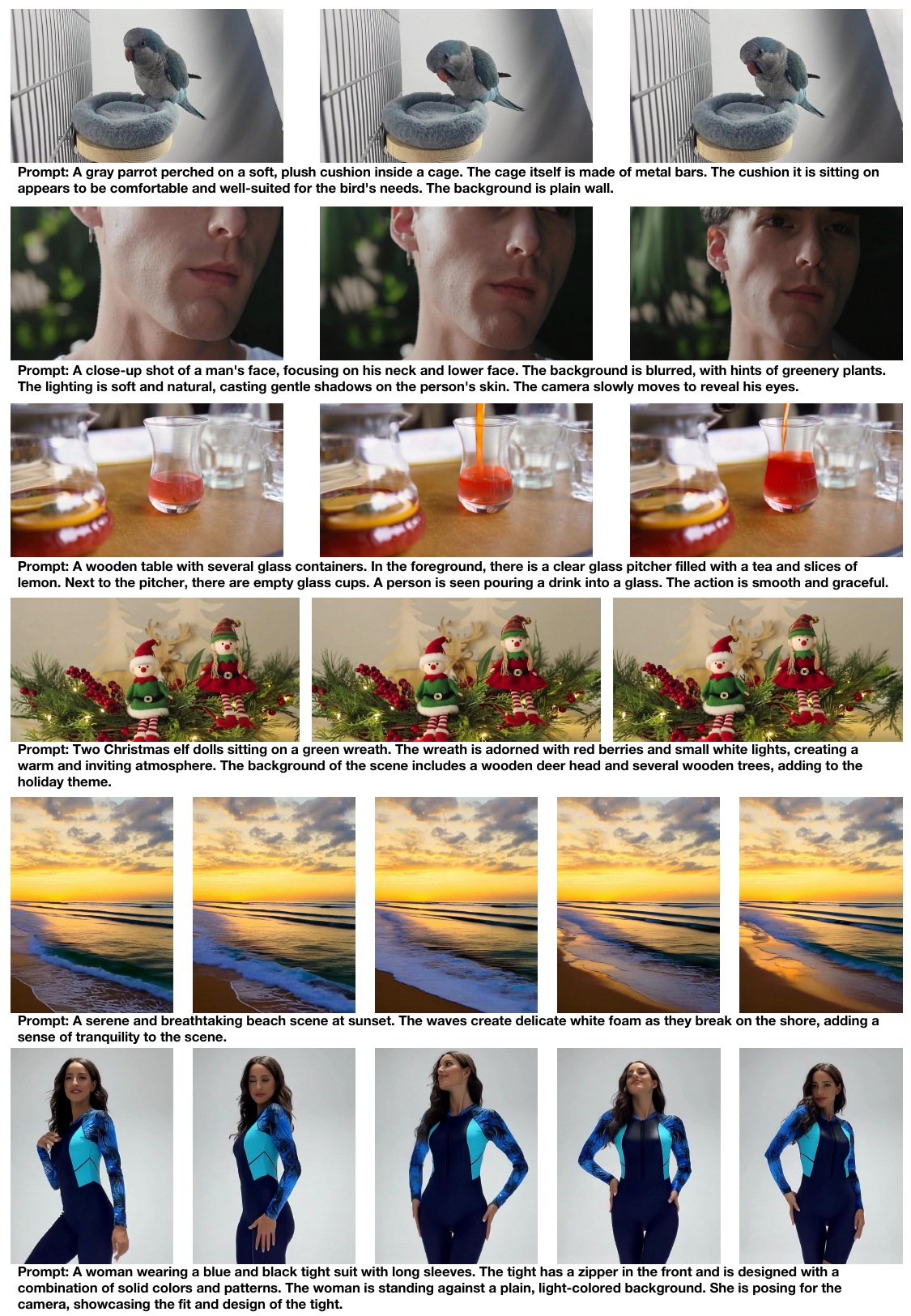} 
    \caption{
    Visualization of text-to-video generation results produced by the MUG-V 10B model. (enlarge for more details.)
    }
    \label{fig:t2v} 
\end{figure}

\begin{figure}
  \centering
\includegraphics[width=0.95\textwidth]{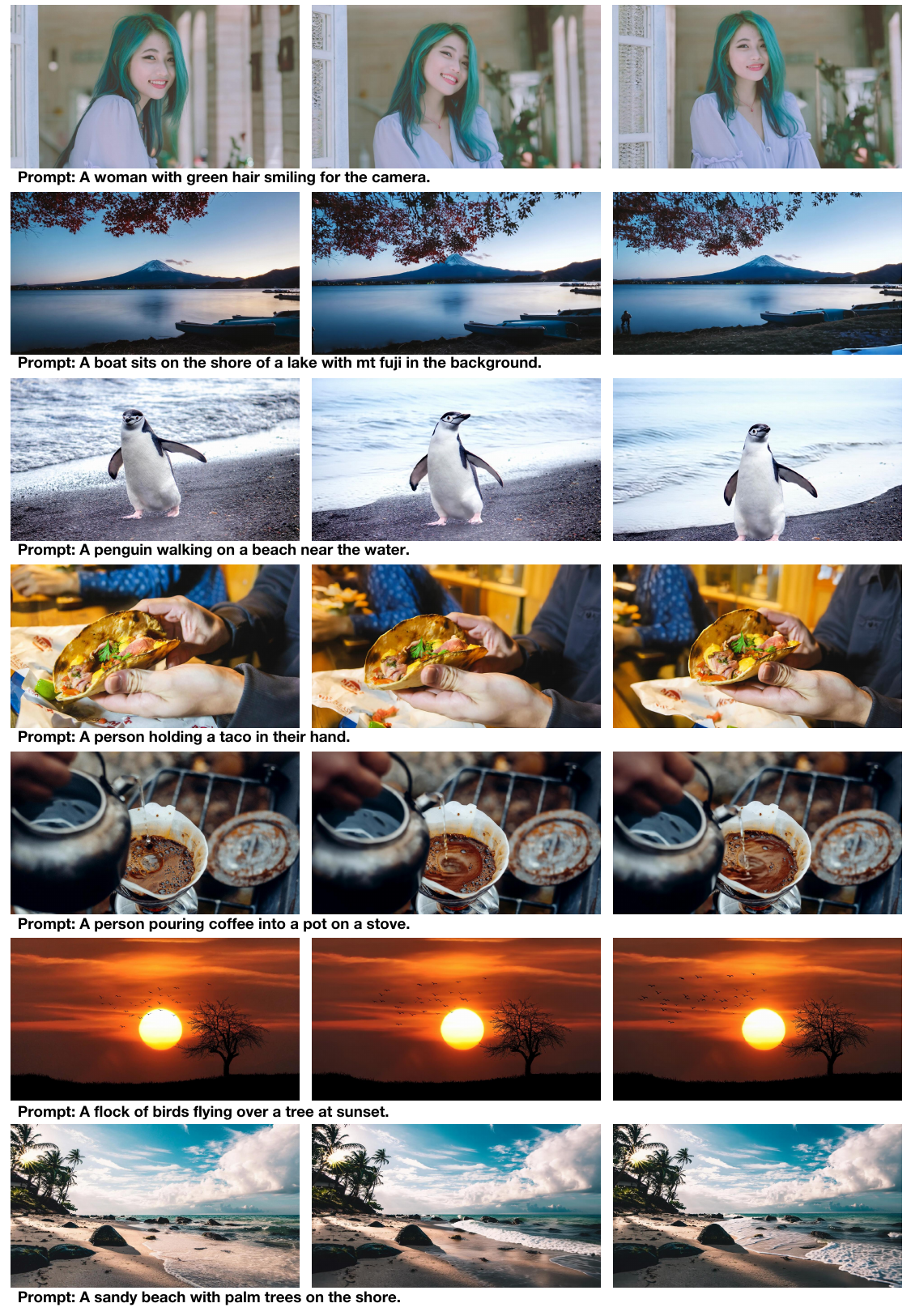} 
    \caption{
    Visualization of image-to-video generation results produced by the MUG-V 10B model. In each example, the first frame corresponds to the conditioning image. (enlarge for more details.)
    }
    \label{fig:i2v} 
\end{figure}

\begin{figure}
  \centering
\includegraphics[width=0.95\textwidth]{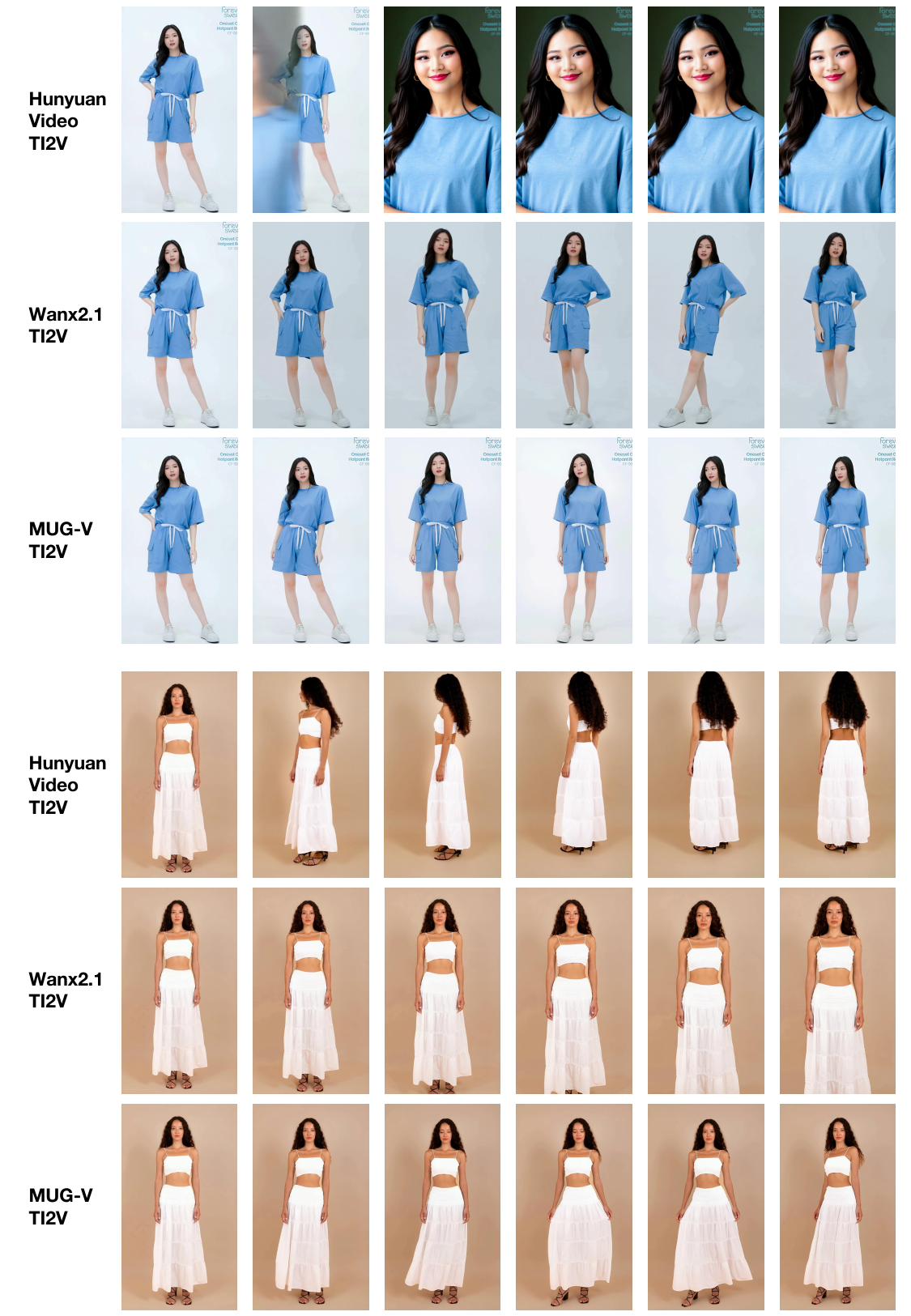} 
    \caption{
    Visualization of e-commerce video generation results across different models. Since each model is optimized for distinct prompt styles, we employed their respective default prompts or prompt-rewriting tools for fair comparison. (enlarge for more details.)
    }
    \label{fig:he_ti2v_1} 
\end{figure}

\begin{figure}
  \centering
\includegraphics[width=0.95\textwidth]{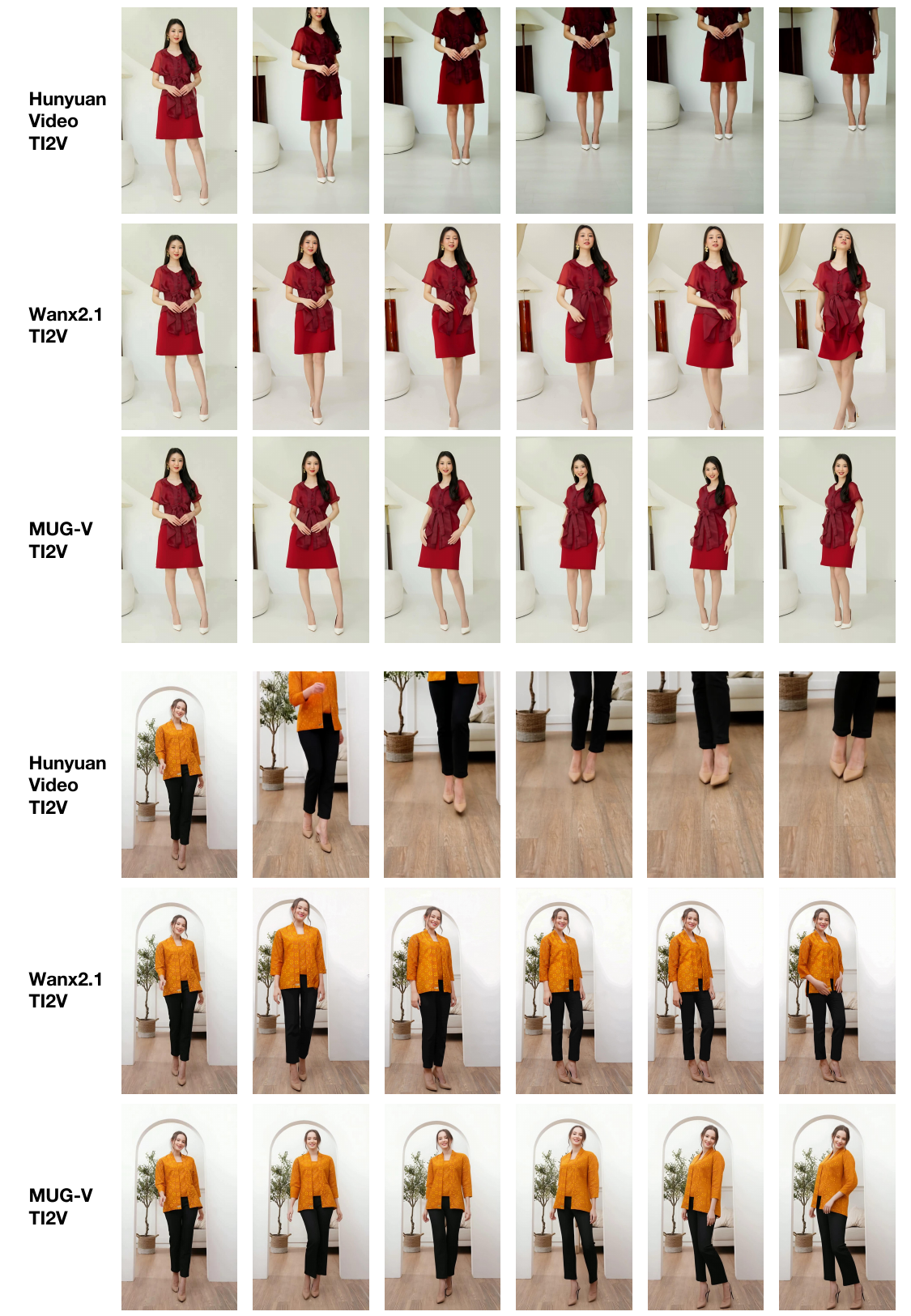} 
    \caption{
    Visualization of e-commerce video generation results across different models. Since each model is optimized for distinct prompt styles, we employed their respective default prompts or prompt-rewriting tools for fair comparison. (enlarge for more details.)
    }
    \label{fig:he_ti2v_2} 
\end{figure}

\begin{figure}
  \centering
\includegraphics[width=0.95\textwidth]{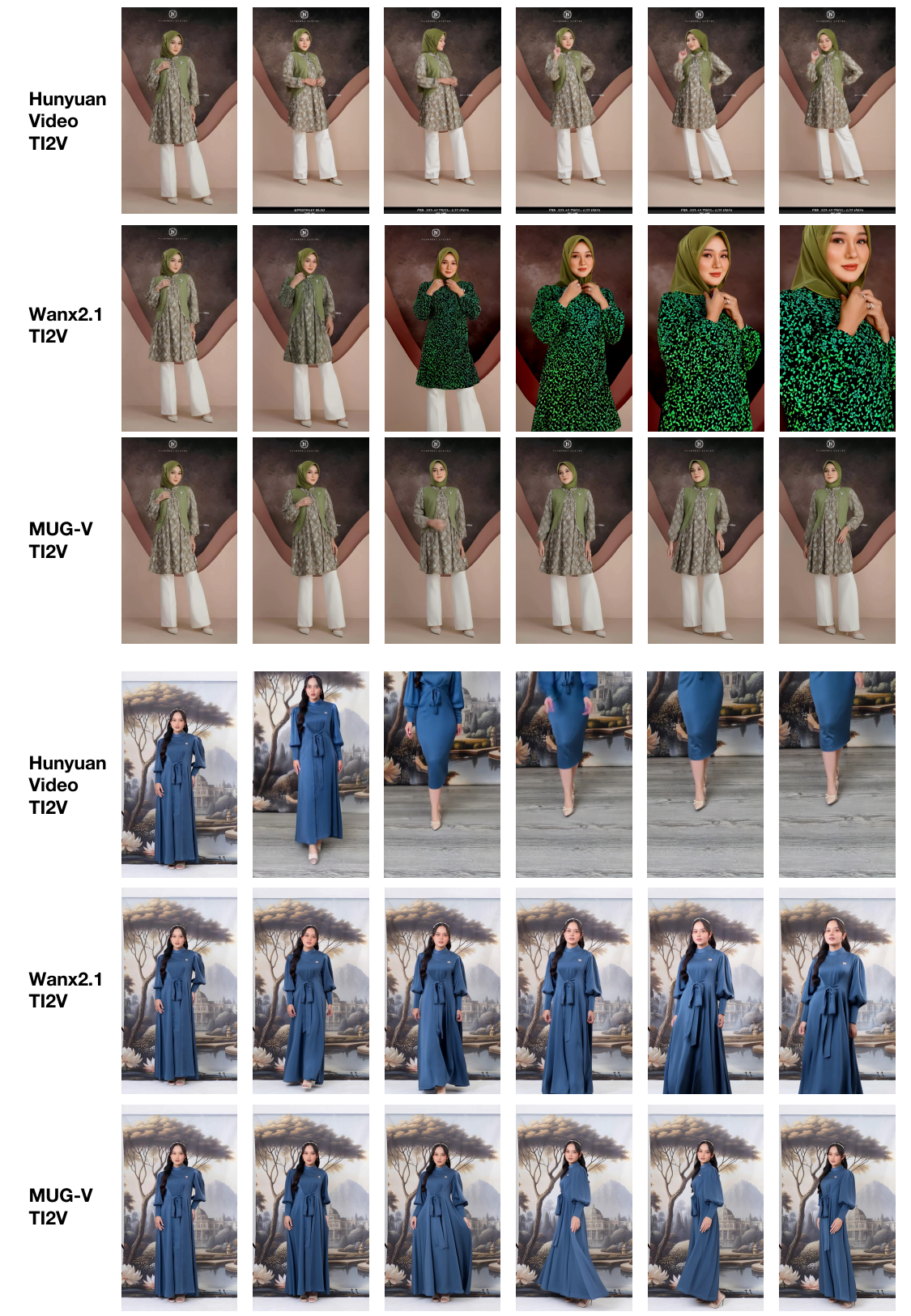} 
    \caption{
    Visualization of e-commerce video generation results across different models. Since each model is optimized for distinct prompt styles, we employed their respective default prompts or prompt-rewriting tools for fair comparison. (enlarge for more details.)
    }
    \label{fig:he_ti2v_3} 
\end{figure}

We present representative qualitative results in Fig.~\ref{fig:t2v} and Fig.~\ref{fig:i2v}. Fig.~\ref{fig:t2v} shows text-to-video (T2V) samples, while Fig.~\ref{fig:i2v} displays image-to-video (I2V) results. Moreover, the generated samples of e-commerce video generation evaluation tasks are presented in Figs.~\ref{fig:he_ti2v_1}, \ref{fig:he_ti2v_2}, \ref{fig:he_ti2v_3}. More video demonstrations are available on our project website.

\section{Challenges and Future Work}
\label{app:chall_and_future}

Despite these advances, our experiments highlight several open challenges. First, strengthening the faithfulness and controllability of the mapping from conditioning signals (text, image, or mixed inputs) to generated videos remains a prerequisite for reliable real-world deployment. Second, fine-grained appearance fidelity, such as material and texture preservation, still lags, with sensitivity to VAE compression and DiT noise initialization leading to subtle but consequential degradations. Third, scaling to longer durations and higher resolutions demands algorithms and systems that cope with long-sequence training, inference efficiency, and long-range temporal consistency. In light of these challenges, we remain committed to advancing the capabilities of video generation models and look forward to continued progress from the broader research community.

\newpage
\bibliographystyle{unsrt}  
\bibliography{references}


\end{document}